\documentclass[10pt,twocolumn,letterpaper]{article}

\usepackage{iccv}
\usepackage{times}
\usepackage{epsfig}
\usepackage{graphicx}
\usepackage{amsmath}
\usepackage{amssymb}

\usepackage{multirow}
\usepackage{xcolor}
\usepackage{epstopdf}
\usepackage{caption}
\usepackage{subcaption}
\usepackage{bm}
\usepackage{footnote}
\usepackage{stfloats}
\usepackage{placeins}
\usepackage{color,soul}
\usepackage{dsfont}
\usepackage{pbox}
\usepackage{enumerate}
\usepackage{etextools}
\usepackage{tabulary}
\usepackage{dsfont}
\usepackage{mathtools}
\usepackage{amssymb}
\usepackage{stfloats}
\usepackage[algoruled,boxed,lined]{algorithm2e}
\usepackage{soul}
\usepackage{epstopdf}
\usepackage{placeins}
\usepackage{color,soul}
\usepackage{dsfont}
\usepackage{amssymb}
\usepackage{pbox}
\usepackage{enumerate}
\usepackage{etextools}
\usepackage{mathtools}
\mathtoolsset{showonlyrefs}
\usepackage{array}

\def\Xb{{\mathbf{\bar{S}}}}
\def\Rb{{\mathbf{\bar{R}}}}
\def\Zb{{\mathbf{\bar{Z}}}}
\def\p{{\mathbf p}}

\def\s{{\mathbf s}}
\def\X{{\mathbf X}}
\def\dd{{\mathbf d}}

\def\u{ \mathbf{u}}

\def\Y{{\mathbf Y}}

\def\Z{{\mathbf Z}}
\def\u{{\mathbf u}}

\def\Q{{\mathbf Q}}

\def\D{{\mathbf D}}

\def\A{{\mathbf A}}

\def\M{{\mathbf{M}}}

\def\R{{\mathbf R}}

\def\V{{\mathbf V}}

\def\alfa{{\boldsymbol \alpha}}
\def\gama{{\boldsymbol \gamma}}
\def\Gama{{\boldsymbol \Gamma}}

\def\Om{{\boldsymbol \Omega}}

\iccvfinalcopy % *** Uncomment this line for the final submission

\def\iccvPaperID{1684} % *** Enter the ICCV Paper ID here

\ificcvfinal\fi
\begin{document}

%%%%%%%%% TITLE
\title{Convolutional Dictionary Learning via Local Processing}

\author{
Vardan Papyan\\
{\tt\small vardanp@campus.technion.ac.il}
\and
Yaniv Romano\\
{\tt\small yromano@tx.technion.ac.il}
\and
Jeremias Sulam\\
{\tt\small jsulam@cs.technion.ac.il}
\and
Michael Elad\\
{\tt\small elad@cs.technion.ac.il}\\
Technion - Israel Institute of Technology\\
Technion City, Haifa 32000, Israel\\
}

\maketitle
%\thispagestyle{empty}

%%%%%%%%% ABSTRACT
\begin{abstract}
Convolutional Sparse Coding (CSC) is an increasingly popular model in the signal and image processing communities, tackling some of the limitations of traditional patch-based sparse representations. Although several works have addressed the dictionary learning problem under this model, these relied on an ADMM formulation in the Fourier domain, losing the sense of locality and the relation to the traditional patch-based sparse pursuit. A recent work suggested a novel theoretical analysis of this global model, providing guarantees that rely on a localized sparsity measure. Herein, we extend this local-global relation by showing how one can efficiently solve the convolutional sparse pursuit problem and train the filters involved, while operating locally on image patches. Our approach provides an intuitive algorithm that can leverage standard techniques from the sparse representations field. The proposed method is fast to train, simple to implement, and flexible enough that it can be easily deployed in a variety of applications. We demonstrate the proposed training scheme for image inpainting and image separation, while achieving state-of-the-art results.
\end{abstract}

%%%%%%%%% BODY TEXT
\section{Introduction}
The celebrated sparse representation model has led to impressive results in various applications over the last decade \cite{elad2006image,aharon2006rm,wright2009robust,yang2010image,Dong2011}. In this context one typically assumes that a signal $\X \in \mathbb{R}^N$ is a linear combination of a few columns, also called atoms, taken from a matrix $\D \in \mathbb{R}^{N \times M}$ termed a dictionary; i.e. $\X = \D \Gama$ where $\Gama \in \mathbb{R}^M$ is a sparse vector. Given $\X$, finding its sparsest representation, called sparse pursuit, amounts to solving the following problem
\begin{equation} \label{Eq:saprse_coding}
\min_{ \Gama } \ \| \Gama \|_0 \ \ \text{s.t.} \ \ \left\| \X - \D \Gama \right\|_2 \leq \epsilon,
\end{equation}
where $\epsilon$ stands for the model mismatch or an additive noise strength.
The solution for the above can be approximated using greedy algorithms such as Orthogonal Matching Pursuit (OMP) \cite{Chen1989} or convex formulations such as BP \cite{Chen2001}. The task of learning the model, i.e. identifying the dictionary $\D$ that best represents a set of training signals, is called dictionary learning and several methods have been proposed for tackling it, including K-SVD \cite{aharon2006rm}, MOD \cite{engan1999method}, online dictionary learning \cite{mairal2009online}, trainlets \cite{sulam2016trainlets}, and more.

\begin{figure*}[t!]
	\centering
	\includegraphics[width=0.8\textwidth]{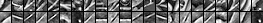}
	\caption{Top: Patches extracted from natural images. Bottom: Their corresponding slices. Observe how the slices are far simpler, and contained by their corresponding patches.}
	\label{Fig:patches_vs_slices}
	\vspace{-.35cm}
\end{figure*}

When dealing with high-dimensional signals, addressing the  dictionary learning problem becomes computationally infeasible, and learning the model suffers from the curse of dimensionality. Traditionally, this problem was circumvented by training a local model for patches extracted from $\X$ and processing these independently. This approach gained much popularity and success due to its simplicity and high-performance \cite{elad2006image,mairal2008sparse,yang2010image,Dong2011,mairal2014sparse}. A different approach is the Convolutional Sparse Coding (CSC) model, which aims to amend the problem by imposing a specific structure on the global dictionary involved \cite{Grosse2007, Bristow2013, Kong2014, Wohlberg2014, Heide2015, Gu2015}. In particular, this model assumes that $\D$ is a banded convolutional dictionary, implying that this global model assumes that the signal is a superposition of a few local atoms, or filters, shifted to different positions. Several works have presented algorithms for training convolutional dictionaries \cite{Bristow2013, Heide2015, Wohlberg2014}, circumventing some of the computational burdens of this problem by relying on ADMM solvers that operate in the Fourier domain. In doing so, these methods lost the connection to the patch-based processing paradigm, as widely practiced in many signal and image processing applications.

In this work, we propose a novel approach for training the CSC model, called slice-based dictionary learning. Unlike current methods, we leverage a localized strategy enabling the solution of the global problem in terms of only local computations in the original domain. The main advantages of our method over existing ones are:
\begin{enumerate}
\item It operates \textit{locally} on patches, while solving faithfully the \textit{global} CSC problem;
\item It reveals how one should modify current (and any) dictionary learning algorithms to solve the CSC problem in a variety of applications;
\item It is easy to implement and intuitive to understand;
\item It can leverage standard techniques from the sparse representations field, such as OMP, LARS, K-SVD, MOD, online dictionary learning and trainlets; 
\item It converges faster than current state of the art methods, while providing a better model; and
\item It can naturally allow for a different number of non-zeros in each spatial location, according to the local signal complexity.
\end{enumerate}

The rest of this paper is organized as follows: Section \ref{Sec:CSC} reviews the CSC model. The proposed method is presented in Section \ref{Sec:proposed_algorithm} and contrasted with conventional approaches in Section \ref{Sec:other_methods}. Section \ref{Sec:image_restoration} shows how our method can be employed to tackle the tasks of image inpainting and separation, and later in Section \ref{Sec:experiments} we demonstrate empirically our algorithms. We conclude this work in Section \ref{Sec:conclusion}.

\section{Convolutional Sparse Coding} \label{Sec:CSC}
The CSC model assumes that a global signal $\X$ can be decomposed as $\X = \sum_{i=1}^m \dd_i \ast \Gama_i$, where $\dd_i \in \mathbb{R}^n$ are local filters that are convolved with their corresponding features maps (or sparse representations) $\Gama_i \in \mathbb{R}^N$. Alternatively, following Figure \ref{Fig:CSC_global_system}, the above can be written in matrix form as $\X = \D \Gama$; where $\D \in \mathbb{R}^{N\times Nm}$ is a banded convolutional dictionary built from shifted versions of a local matrix $\D_L$, containing the atoms $\{ \dd_i \}_{i=1}^m$ as its columns, and $\Gama \in \mathbb{R}^{Nm}$ is a global sparse representation obtained by interlacing the $\{ \Gama_i \}_{i=1}^m$. In this setting, a patch $\R_i \X$ taken from the global signal equals $\Om \gama_i$, where $\Om \in \mathbb{R}^{n \times (2n-1)m}$ is a \textit{stripe} dictionary and $\gama_i \in \mathbb{R}^{(2n-1)m}$ is a \textit{stripe} vector. Here we defined $\R_i \in \mathbb{R}^{n \times N}$ to be the operator that extracts the $i$-th $n$-dimensional patch from $\X$.

\begin{figure}[t!]
	\centering
	\includegraphics[width=0.5\textwidth]{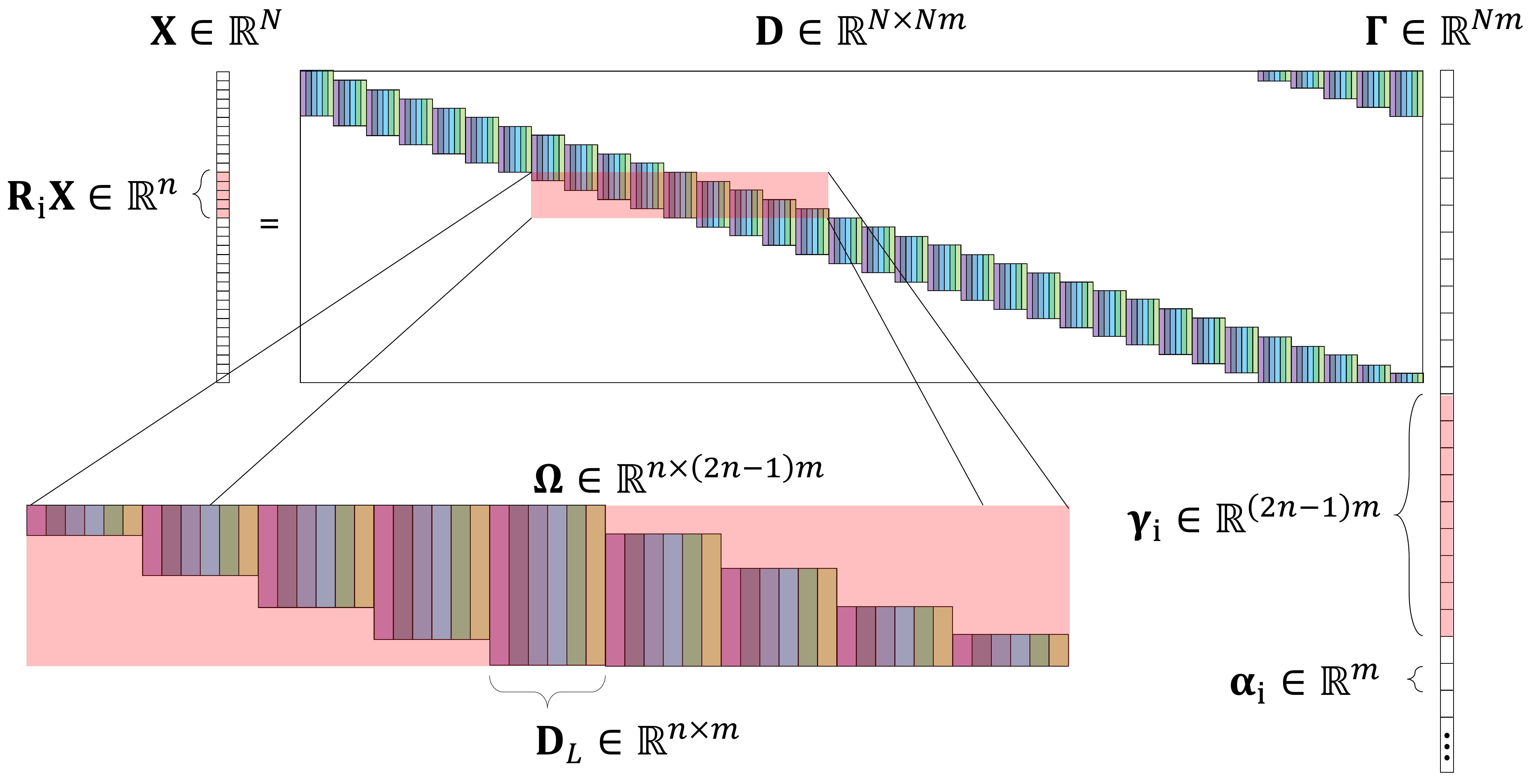}
	\caption{The CSC model and its constituent elements.}
	\label{Fig:CSC_global_system}
\end{figure}

The work in \cite{Papyan2016_1} suggested a theoretical analysis of this global model, driven by a localized sparsity measure. Therein, it was shown that if all the stripes $\gama_i$ are sparse, the solution to the convolutional sparse pursuit problem is unique and can be recovered by greedy algorithms, such as the OMP \cite{Chen1989}, or convex formulations such as the Basis Pursuit (BP) \cite{Chen2001}. This analysis was then extended in \cite{Papyan2016_2} to a noisy regime showing that, under similar sparsity assumptions, the global problem formulation and the pursuit algorithms are also stable. Herein, we leverage this local-global relation from an algorithmic perspective, showing how one can efficiently solve the convolutional sparse pursuit problem and train the dictionary (i.e., the filters) involved, while only operating locally.

Note that the global sparse vector $\Gama$ can be broken into a set of non-overlapping $m$-dimensional sparse vectors $\alfa_{i=1}^N$, which we call \textit{needles}. The essence of the presented algorithm is in the observation that one can express the global signal as $\X = \sum_{i=1}^N \R_i^T \D_L \alfa_i$, where $\R_i^T \in \mathbb{R}^{N \times n}$ is the operator that puts $\D_L \alfa_i$ in the $i$-th position and pads the rest of the entries with zeros. Denoting by $\s_i$ the $i$-th \textit{slice} $\D \alfa_i$, we can write the above as $\X = \sum_{i=1}^N \R_i^T \s_i$. It is important to stress that the slices do not correspond to patches extracted from the signal, $\R_i \X$, but rather to much simpler entities. They represent only a fraction of the $i$-th patch, since $\R_i \X = \R_i \sum_{j=1}^N \R_j^T \s_j$, i.e. a patch is constructed from several overlapping slices. Unlike current works in signal and image processing, which train a local dictionary on the patches $\{ \R_i \X \}_{i=1}^N$, in what follows we define the learning problem with respect to the slices, $\{ \s_i \}_{i=1}^N$, instead. In other words, we aim to train $\D_L$ instead of $\Om$. As a motivation, we present in Figure \ref{Fig:patches_vs_slices} a set of patches $\R_i \X$ extracted from natural images and their corresponding slices $\s_i$, obtained from the proposed algorithm, which will be presented in Section \ref{Sec:proposed_algorithm}. Indeed, one can observe that the slices are simpler than the patches, as they contain less information.

\section{Proposed Method: Slice-Based Dictionary Learning} \label{Sec:proposed_algorithm}
The convolutional dictionary learning problem refers to the following optimization\footnote{Hereafter, we assume that the atoms in the dictionary are normalized to a unit $\ell_2$ norm.} objective,
\begin{equation} \label{Eq:conv_SC_DL}
	\min_{\D,\Gama} \frac{1}{2} \| \X - \D\Gama \|_2^2 + \lambda \| \Gama \|_1,
\end{equation}
for a convolutional dictionary $\D$ as in Figure \ref{Fig:CSC_global_system} and a Lagrangian parameter $\lambda$ that controls the sparsity level. Employing the decomposition of $\X$ in terms of its slices, and the separability of the $\ell_1$ norm, the above can be written as the following constrained minimization problem,
\begin{align}
	\min_{\D_L, \{ \alfa_i \}_{i=1}^N, \{ \s_i \}_{i=1}^N } & \frac{1}{2} \| \X - \sum_{i=1}^N \R_i^T \s_i \|_2^2 + \lambda \sum_{i=1}^N \| \alfa_i \|_1 \\
	& \text{s.t.} \ \ \s_i = \D_L \alfa_i.
\end{align}
One could tackle this problem using half-quadratic splitting \cite{geman1995nonlinear} by introducing a penalty term over the violation of the constraint and gradually increasing its importance. Alternatively, we can employ the ADMM algorithm \cite{Boyd2011} and solve the augmented Lagrangian formulation (in its scaled form),
\begin{align} \label{Eq:single_layer_ADMM}
	\min_{ \substack{ \D_L, \{ \alfa_i \}_{i=1}^N, \\ \{ \s_i \}_{i=1}^N, \{\u_i\}_{i=1}^N } } & \frac{1}{2} \| \X - \sum_{i=1}^N \R_i^T \s_i \|_2^2 \\
	+ & \sum_{i=1}^N \left( \lambda \| \alfa_i \|_1 + \frac{\rho}{2} \| \s_i - \D_L \alfa_i + \u_i \|_2^2 \right),
\end{align}
where $\{ \u_i \}_{i=1}^N$ are the dual variables that enable the constrains to be met.

\subsection{Local Sparse Coding and Dictionary Update}
The minimization of Equation \eqref{Eq:single_layer_ADMM} with respect to all the needles $\{ \alfa_i \}_{i=1}^N$ is separable, and can be addressed independently for every $\alfa_i$ by leveraging standard tools such as LARS. This also allows for having a different number of non-zeros per slice, depending on the local complexity. Similarly, the minimization with respect to $\D_L$ can be done using any patch-based dictionary learning algorithm such as the K-SVD, MOD, online dictionary learning or trainlets. Note that in the dictionary update stage, while minimizing for $\D_L$ and $\{\alfa_i\}$, one could refrain from iterating these updates until convergence, and instead perform only a few iterations before proceeding with the remaining variables.
	
\subsection{Slice Update via Local Laplacian} \label{Subsection:slice_update}
The minimization of Equation \eqref{Eq:single_layer_ADMM} with respect to all the slices $\{ \s_i \}_{i=1}^N$ amounts to solving the following quadratic problem
\begin{equation}
	\min_{ \{ \s_i \}_{i=1}^N } \frac{1}{2} \| \X - \sum_{i=1}^N \R_i^T \s_i \|_2^2 + \frac{\rho}{2} \sum_{i=1}^N \| \s_i - \D_L \alfa_i + \u_i \|_2^2.
\end{equation}
Taking the derivative with respect to the variables $\s_1, \s_2, \dots \s_N$ and nulling them, we obtain the following system of linear equations
\begin{align}
	& \R_1 ( \sum_{i=1}^N \R_i^T \s_i - \X ) + \rho (\s_1 - \D_L \alfa_1 + \u_1) = \mathbf{0} \\
	& \phantom{................................} \vdots \\
	& \R_N ( \sum_{i=1}^N \R_i^T \s_i - \X ) + \rho (\s_N - \D_L \alfa_N + \u_N) = \mathbf{0}.
\end{align}
Defining
\begin{align}
	\Rb =
	\begin{bmatrix}
		\R_1 \\
		\R_2 \\
		\vdots \\
		\R_N
	\end{bmatrix}
	\quad
	\Xb = \begin{bmatrix}
		\s_1 \\
		\s_2 \\
		\vdots \\
		\s_N \\
	\end{bmatrix}
	\quad
	\Zb =
	\begin{bmatrix}
		\D_L \alfa_1 - \u_1 \\
		\D_L \alfa_2 - \u_2 \\
		\vdots \\
		\D_L \alfa_N - \u_N \\
	\end{bmatrix},
\end{align}
the above can be written as
\begin{align}
	& \mathbf{0} = \Rb \left( \Rb^T \Xb - \X \right) + \rho \left( \Xb - \Zb \right) \\
	& \implies \Xb = \left( \Rb \Rb^T  + \rho \mathbf{I} \right)^{-1} \left( \Rb \X + \rho \Zb \right).
\end{align}
Using the Woodbury matrix identity and the fact that $\Rb^T \Rb = \sum_{i=1}^N \R_i^T \R_i = n \mathbf{I}$, where $\mathbf{I}$ is the identity matrix, the above is equal to
\begin{align}
	\Xb & = \left( \frac{1}{\rho} \mathbf{I} - \frac{1}{\rho^2} \Rb \left( \mathbf{I} + \frac{1}{\rho} \Rb^T \Rb \right)^{-1} \Rb^T \right) \left( \Rb \X + \rho \Zb \right) \\
	& = \left( \frac{1}{\rho} \mathbf{I} - \frac{1}{\rho^2} \Rb \left( \mathbf{I} + \frac{1}{\rho} n \mathbf{I} \right)^{-1} \Rb^T \right) \left( \Rb \X + \rho \Zb \right) \\
	& = \left( \mathbf{I} - \Rb \left( \rho \mathbf{I} + n \mathbf{I} \right)^{-1} \Rb^T \right) \left( \frac{1}{\rho} \Rb \X + \Zb \right).
\end{align}		
Plugging the definitions of $\Rb$, $\Xb$ and $\Zb$, we obtain
\begin{align} \label{Eq:single_layer_local_laplacian}
	\s_i = & \left( \frac{1}{\rho} \R_i \X + \D_L \alfa_i - \u_i \right) \\
	- \R_i & \left( \frac{1}{\rho + n} \sum_{j=1}^N \R_j^T \left( \frac{1}{\rho} \R_j \X + \D_L \alfa_j - \u_j \right) \right).
\end{align}
Although seemingly complicated at first glance, the above is simple to interpret and implement in practice. This expression indicates that one should (i) compute the estimated slices $\p_i = \frac{1}{\rho} \R_i \X + \D_L \alfa_i - \u_i$, then (ii) aggregate them to obtain the global estimate $\hat{\X} = \sum_{j=1}^N \R_j^T \p_j$, and finally (iii) subtract from $\p_i$ the corresponding patch from the aggregated signal, i.e. $\R_i \hat{\X}$. As a remark, since this update essentially subtracts from $\p_i$ an averaged version of it, it can be seen as some sort of a patch-based local Laplacian operation.

\begin{algorithm}[t]
	\SetKwInOut{Input}{Input}
	\SetKwInOut{Output}{Output}
	
	\Input{Signal $\X$, initial dictionary $\D_L$}
	\Output{Trained dictionary $\D_L$, needles $\{ \alfa_i \}_{i=1}^N$ and slices $\{ \s_i \}_{i=1}^N$}
	
	\textbf{Initialization:} \vspace{-0.3cm}
	\begin{equation}
	\s_i = \frac{1}{n} \R_i \X, \quad \u_i = \mathbf{0}
	\end{equation}
	
    \vspace{-0.1cm}

	\For{$iteration = 1 : T$}{
	
		\vspace{0.25cm}
		
		Local sparse pursuit (needle): \vspace{-0.3cm}
		\begin{equation}
		\alfa_i = \underset{\alfa_i}{\arg\min} \ \frac{\rho}{2} \| \s_i - \D_L \alfa_i + \u_i \|_2^2 + \lambda \| \alfa_i \|_1
		\end{equation}
		
		\vspace{-0.3cm}
		
		Slice reconstruction: \vspace{-0.3cm}
		\begin{equation}
		\p_i = \frac{1}{\rho} \R_i \X + \D_L \alfa_i - \u_i
		\end{equation}
		
		\vspace{-0.3cm}
		
		Slice aggregation: \vspace{-0.5cm}
		\begin{equation}
		\hat{\X} = \sum_{j=1}^N \R_j^T \p_j
		\end{equation}
		
		\vspace{-0.3cm}
		
		Slice update via local Laplacian: \vspace{-0.3cm}
		\begin{equation}
		\s_i = \p_i - \frac{1}{\rho + n} \R_i \hat{\X}
		\end{equation}

	 	\vspace{-0.3cm}
	 
		Dual variable update: \vspace{-0.35cm}
		\begin{equation}
		\u_i = \u_i + \s_i - \D_L\alfa_i
		\end{equation}
		
		\vspace{-0.3cm}
		 
		Dictionary update: \vspace{-0.5cm}
		\begin{equation}
		\D_L = \underset{\D_L, \{ \alfa_i \}_{i=1}^N}{\arg\min} \ \sum_{i=1}^N \| \s_i - \D_L \alfa_i + \u_i \|_2^2
		\end{equation}
		
		 \vspace{-0.4cm}
	}
	\caption{Slice-based dictionary learning} \label{Alg:single_layer}
\end{algorithm}

\subsection{Boundary Conditions} \label{Sec:boundaries}
In the description of the CSC model (see Figure \ref{Fig:CSC_global_system}), we assumed for simplicity circulant boundary conditions. In practice, however, natural signals such as images are in general not circulant and special treatment is needed for the boundaries. One way of handling this issue is by assuming that $\X = \M \D \Gama$, where $\M \in \mathbb{R}^{N \times N+2(n-1)}$ is matrix that crops the first and last $n-1$ rows of the dictionary $\D$ (see Figure \ref{Fig:CSC_global_system}). The change needed in Algorithm \ref{Alg:single_layer} to incorporate $\M$ is minor. Indeed, one has to simply replace the patch extraction operator $\R_i$, with $\R_i \M^T$, where the operator $\M^T \in \mathbb{R}^{N+2(n-1) \times N}$ pads a global signal with $n-1$ zeros on the boundary and $\R_i$ extracts a patch from the result. In addition, one has to replace the patch placement operator $\R_i^T$ with $\M \R_i^T$, which simply puts the input in the location of the $i$-th patch and then crops the result.

\subsection{From Patches to Slices}
The ADMM variant of the proposed algorithm, named \textit{slice-based dictionary learning}, is summarized in Algorithm \ref{Alg:single_layer}. While we have assumed the data corresponds to one signal $\X$, this can be easily extended to consider several signals.

\begin{figure}[b!]
	\centering
	\begin{subfigure}{0.4\textwidth}
		\centering
		\includegraphics[width=1\textwidth]{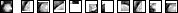}
	\end{subfigure}
	\\[0.1cm]
	\begin{subfigure}{0.4\textwidth}
		\centering
		\includegraphics[width=1\textwidth]{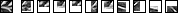}
	\end{subfigure}
	\\[0.1cm]
	\begin{subfigure}{0.4\textwidth}
		\centering
		\includegraphics[width=1\textwidth]{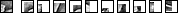}
	\end{subfigure}
	\\[0.1cm]
	\begin{subfigure}{0.4\textwidth}
		\centering
		\includegraphics[width=1\textwidth]{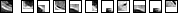}
	\end{subfigure}
	\\[0.1cm]
	\begin{subfigure}{0.4\textwidth}
		\centering
		\includegraphics[width=1\textwidth]{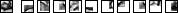}
	\end{subfigure}
	\\[0.1cm]
	\begin{subfigure}{0.4\textwidth}
		\centering
		\includegraphics[width=1\textwidth]{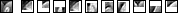}
	\end{subfigure}
	\\[0.1cm]
	\begin{subfigure}{0.4\textwidth}
		\centering
		\includegraphics[width=1\textwidth]{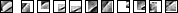}
	\end{subfigure}
	\\[0.1cm]
	\begin{subfigure}{0.4\textwidth}
		\centering
		\includegraphics[width=1\textwidth]{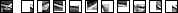}
	\end{subfigure}
	\\[0.1cm]
	\begin{subfigure}{0.4\textwidth}
		\centering
		\includegraphics[width=1\textwidth]{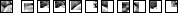}
	\end{subfigure}
	\\[0.1cm]
	\begin{subfigure}{0.4\textwidth}
		\centering
		\includegraphics[width=1\textwidth]{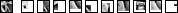}
	\end{subfigure}
	\\[0.1cm]
	\begin{subfigure}{0.4\textwidth}
		\centering
		\includegraphics[width=1\textwidth]{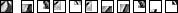}
	\end{subfigure}
	\caption{The first column contains patches extracted from the training data, and second to eleventh columns are the corresponding slices constructing these patches. For each patch, only the ten slices with the highest energy are presented.}
	\label{Fig:patch_decomposition}
\end{figure}

At this point, a discussion regarding the relation between this algorithm and standard (patch-based) dictionary learning techniques is in place. Indeed, from a quick glance the two approaches seem very similar: Both perform local sparse pursuit on local patches extracted from the signal, then update the dictionary to represent these patches better, and finally apply patch-averaging to obtain a global estimate of the reconstructed signal. Moreover, both iterate this process in a block-coordinate descent manner in order to minimize the overall objective. So, what is the difference between this algorithm and previous approaches?

The answer lies in the \textit{migration from patches to slices}. While originally dictionary learning algorithms aimed to represent patches $\R_i \X$ taken from the signal, our scheme suggests to train the dictionary to construct slices, which do not necessarily reconstruct the patch fully. Instead, only the summation of these slices results in the reconstructed patches. To illustrate this relation, we show in Figure \ref{Fig:patch_decomposition} the decomposition of several patches in terms of their constituent slices. One can observe that although the slices are simple in nature, they manage to construct the rather complex patches. The difference between this illustration and that of Figure \ref{Fig:patches_vs_slices} is that the latter shows patches $\R_i \X$ and only the slices that are \emph{fully} contained in them.

Note that the slices are not mere auxiliary variables, but rather emerge naturally from the convolutional formulation. After initializing these with patches from the signal, $\s_i = \frac{1}{n} \R_i \X$, each iteration progressively ``carves'' portions from the patch via the local Laplacian, resulting in simpler constructions. Eventually, these variables are guaranteed to converge to $\D_L \alfa_i$ -- the slices we have defined.

Having established the similarities and differences between the traditional patch-based approach and the slice alternative, one might wonder what is the advantage of working with slices over patches. In the conventional approach, the patches are processed independently, ignoring their overlap. In the slice-based case, however, the local Laplacian forces the slices to communicate and reach a consensus on the reconstructed signal. Put differently, the CSC offers a global model, while earlier patch-based methods used local models without any holistic fusion of them.

\section{Comparison to Other Methods} \label{Sec:other_methods}
In this section we explain further the advantages of our method, and compare it to standard algorithms for training the CSC model such as \cite{Heide2015,wohlberg2016boundary}. Arguably the main difference resides in our localized treatment, as opposed to the global Fourier domain processing. Our approach enables the following benefits:
\begin{enumerate}
\item The sparse pursuit step can be done separately for each slice and is therefore trivial to parallelize.
\item The algorithm can work in a complete online regime where in each iteration it samples a random subset of slices, solves a pursuit for these and then updates the dictionary accordingly. Adopting a similar strategy in the competing algorithms \cite{Heide2015,wohlberg2016boundary} might be problematic, since these are deployed in the Fourier domain on global signals and it is therefore unclear how to operate on a subset of local patches.
\item Our algorithm can be easily modified to allow a different number of non-zeros in each location of the global signal. Such local adaptation to the complexity of the image cannot be offered by the Fourier-oriented algorithms. 
\end{enumerate}

We now turn to comparing the proposed algorithm to alternative methods in terms of computational complexity. Denote by $I$ the number of signals on which the dictionary is trained, and by $k$ the maximal number of non-zeros in a needle\footnote{Although we solve the Lagrangian formulation of LARS, we also limit the maximal number of non-zeros per needle to be at most $k$.} $\alfa_i$. At each iteration of our algorithm we employ LARS that has a complexity of $O( k^3 + m k^2 + n m )$ per slice \cite{mairal2014sparse}, resulting in $O( IN(k^3 + m k^2 + n m) + n m^2 )$ computations for all $N$ slices in all the $I$ images. The last term, $n m^2$, corresponds to the precomputation of the Gram of the dictionary $\D_L$ (which is in general negligible). Then, given the obtained needles, we reconstruct the slices, requiring $O( IN n k )$, aggregate the results to form the global estimate, incurring $O( IN n )$, and update the slices, which requires an additional $O( IN n )$. These steps are negligible compared to the sparse pursuits and are thus omitted in the final expression. Finally, we update the dictionary using the K-SVD, which is $O( n m^2 + I N k n + I N k m )$ \cite{rubinstein2008efficient}. We summarize the above in Table \ref{Table:complexity_analysis}. In addition, we present in the same table the complexity of each iteration of the (Fourier-based) algorithm in \cite{Heide2015}. In this case, $q$ corresponds to the number of inner iterations in their ADMM solver of the sparse pursuit and dictionary update.

The most computationally demanding step in our algorithm is the local sparse pursuit, which is $O( N I ( k^3 + m k^2 + n m) )$. Assuming that the needles are very sparse, which indeed happens in all of our experiments, this reduces to $O( N I m n )$. On the other hand, the complexity in the algorithm of \cite{Heide2015} is dominated by the computation of the FFT, which is $O( N I m q \log (N) )$. We conclude that our algorithm scales linearly with the global dimension, while theirs grows as $N \log (N)$. Note that this also holds for other related methods, such as that of \cite{wohlberg2016boundary}, which also depend on the global FFT. Moreover, one should remember the fact that in our scheme one might run the pursuits on a small percentage of the total number of slices, meaning that in practice our algorithm can scale as $O( \mu N I n m )$, where $\mu$ is a constant smaller than one.

\begin{table}[t]
	\centering
	\begin{tabular}{|p{0.04\textwidth}|c|}
		\hline
		{\scriptsize Method}  & {\scriptsize Time Complexity}                                                 \\ \hline
		{\scriptsize \begin{tabular}[c]{@{}c@{}} \cite{Heide2015} \\ $I < m$\end{tabular} } & {\scriptsize $ \underbrace{ m I^2 N + (q-1) m I N }_{\text{linear systems}} + \underbrace{ {\color{red} \boldsymbol{q} \boldsymbol{I} \boldsymbol{m} \boldsymbol{N} \boldsymbol{\log (N)}} }_{\text{FFT}} + \underbrace{ q I mN }_{\text{thresholding}} $} \\ \hline
		{\scriptsize \begin{tabular}[c]{@{}c@{}} \cite{Heide2015} \\ $I \geq m$\end{tabular} } & {\scriptsize $ \underbrace{ m^3 N + (q-1) m^2 N }_{\text{linear systems}} + \underbrace{ {\color{red} \boldsymbol{q} \boldsymbol{I} \boldsymbol{m} \boldsymbol{N} \boldsymbol{\log (N)}} }_{\text{FFT}} + \underbrace{ q I mN }_{\text{thresholding}} $} \\ \hline
		{\scriptsize Ours} & {\scriptsize $\underbrace{ {\color{red}\boldsymbol{I} \boldsymbol{N} \boldsymbol{n} \boldsymbol{m}} + I N ( k^3 + m k^2 )}_{\text{LARS / OMP}} + \underbrace{ n m^2 }_{\text{Gram}} + \underbrace{ I N k (n + m) + n m^2 }_{\text{K-SVD}}$ }\\ \hline
	\end{tabular}
	\caption{Complexity analysis. For the convenience of the reader, the dominant term is highlighted in red color.}
	\label{Table:complexity_analysis}
	\vspace{-0.5cm}
\end{table}

\section{Image Processing via CSC} \label{Sec:image_restoration}
In this section, we demonstrate our proposed algorithm on several image processing tasks. Note that the discussion thus far focused on one dimensional signals, however it can be easily generalized to images by replacing the convolutional structure in the CSC model with block-circulant circulant-block (BCCB) matrices.

\subsection{Image Inpainting} \label{Sec:inpainting}
Assume an original image $\X$ is multiplied by a diagonal binary matrix $\A \in \mathbb{R}^{N \times N}$, which masks the entries $\X_i$ in which $\A(i,i) = 0$. In the task of image inpainting, given the corrupted image $\Y = \A \X$, the goal is to restore the original unknown $\X$. One can tackle this problem by solving the following CSC problem
\begin{equation}
\min_{\Gama} \frac{1}{2} \| \Y - \A \D\Gama \|_2^2 + \lambda \| \Gama \|_1,
\end{equation}
where we assume the dictionary $\D$ was pretrained. Using similar steps to those leading to Equation \eqref{Eq:single_layer_ADMM}, the above can be written as
\begin{align}
\min_{ \substack{ \{ \alfa_i \}_{i=1}^N, \{ \s_i \}_{i=1}^N, \\ \{ \u_i \}_{i=1}^N } } & \frac{1}{2} \| \Y - \A \sum_{i=1}^N \R_i^T \s_i \|_2^2 \\
+ & \sum_{i=1}^N \left( \lambda \| \alfa_i \|_1 + \frac{\rho}{2} \| \s_i - \D_L \alfa_i + \u_i \|_2^2 \right).
\end{align}
This objective can be minimized via the algorithm described in the previous section. Moreover, the minimization with respect to the local sparse codes $\{ \alfa_i \}_{i=1}^N$ remains the same. The only difference regards the update of the slices $\{ \s_i \}_{i=1}^N$, in which case one obtains the following expression
\begin{align}
\s_i = & \left( \frac{1}{\rho} \R_i \Y + \D_L \alfa_i - \u_i \right) \\
- & \R_i \left( \frac{1}{\rho + n} \A \sum_{j=1}^N \R_j^T \left( \frac{1}{\rho} \R_j \Y + \D_L \alfa_j - \u_j \right) \right).
\end{align}
The steps leading to the above equation are almost identical to those in subsection \ref{Subsection:slice_update}, and they only differ in the incorporation of the mask $\A$.

\subsection{Texture and Cartoon Separation} \label{Sec:separation}
In this task the goal is to decompose an image $\X$ into its texture component $\X_T$ that contains highly oscillating or pseudo-random patterns, and a cartoon part $\X_C$ that is a piece-wise smooth image. Many image separation algorithms tackle this problem by imposing a prior on both components. For cartoon, one usually employs the isotropic (or anisotropic) Total Variation norm, denoted by $\| \X_C \|_{TV}$. The modeling of texture, on the other hand, is more difficult and several approaches have been considered over the years \cite{elad2005simultaneous,aujol2006structure,ono2014cartoon,zhangconvolutional}.

In this work, we propose to model the texture component using the CSC model. As such, the task of separation amounts to solving the following problem
\begin{align}
\min_{ { \D_T, \Gama_T, \X_C } } & \frac{1}{2} \left\| \X - \D_T \Gama_T - \X_C \right\|_2^2 + \lambda \left\| \Gama_T \right\|_1 + \xi \| \X_C \|_{TV},
\end{align}
where $\D_T$ is a convolutional (texture) dictionary, and $\Gama_T$ is its corresponding sparse vector. Using similar derivations to those presented in Section \ref{Subsection:slice_update}, the above is equivalent to
\begin{align}
\min_{ \substack{ \D_L, \alfa_T^i, \s_T^i, \\ \X_C, \Z_C } } & \frac{1}{2} \left\| \X - \sum_{i=1}^N \R_i^T \s_T^i - \X_C \right\|_2^2 \\
+ & \lambda \sum_{i=1}^N \left\| \alfa_T^i \right\|_1 + \xi \| \Z_C \|_{TV} \\
& \text{s.t.} \quad \s_T^i = \D_L \alfa_T^i, \quad \X_C = \Z_C,
\end{align}
where we split the variable $\X_C$ into $\X_C = \Z_C$ in order to facilitate the minimization over the TV norm. Its corresponding ADMM formulation\footnote{Disregarding the training of the dictionary, this is a standard two-function ADMM problem. The first set of variables are $\{ \s_T^i \}_{i=1}^N$ and $\X_C$, and the second are $\{ \alfa_T^i \}_{i=1}^N$ and $\Z_C$.} is given by
\begin{align}
\min_{ \substack{ \D_L, \alfa_T^i, \s_T^i, \u_T^i, \\ \X_C, \Z_C, \V_C } } & \frac{1}{2} \left\| \X - \sum_{i=1}^N \R_i^T \s_T^i - \X_C \right\|_2^2 \\
+ & \sum_{i=1}^N \left( \frac{\rho}{2} \left\| \s_T^i - \D_L \alfa_T^i + \u_T^i \right\|_2^2 + \lambda \left\| \alfa_T^i \right\|_1 \right) \\
+ & \frac{\eta}{2} \left\| \X_C - \Z_C + \V_C \right\|_2^2 + \xi \| \Z_C \|_{TV},
\end{align}
where $\{ \s_T^i \}_{i=1}^N$, $\{ \alfa_T^i \}_{i=1}^N$ and $\{ \u_T^i \}_{i=1}^N$ are the texture slices, needles and dual variables, respectively, and $\V_C$ is the dual variable of the global cartoon $\X_C$. The above optimization problem can be minimized by slightly modifying Algorithm \ref{Alg:single_layer}. The update for $\{ \alfa_i \}_{i=1}^N$ is a sparse pursuit and the update for the $\Z_C$ variable is a TV denoising problem. Then, one can update the $\{ \s_T^i \}_{i=1}^N$ and $\X_C$ jointly by
\begin{align}
\s_T^i & = \frac{1}{\rho} \p_T^i - \frac{ \frac{1}{\rho} }{ 1 + \frac{n^2}{\rho} + \frac{1}{\eta} } \R_i \left( \frac{1}{\rho} \sum_{j=1}^N \R_j^T \p_T^j + \frac{1}{\eta} \Q_C \right) \\
\X_C & = \frac{1}{\eta} \Q_C - \frac{ \frac{1}{\eta} }{ 1 + \frac{n^2}{\rho} + \frac{1}{\eta} } \left( \frac{1}{\rho} \sum_{j=1}^N \R_j^T \p_T^j + \frac{1}{\eta} \Q_C \right),
\end{align}
where
$\p_T^i = \R_i \X + \rho \left( \D_L \alfa_T^i - \u_T^i \right)$ and $\Q_C = \X + \eta \left( \Z_C - \V_C \right)$. The final step of the algorithm is updating the texture dictionary $\D_L$ via any dictionary learning method.

\begin{figure}[b]
	\centering
	\begin{subfigure}{0.205\textwidth}
		\centering
		\includegraphics[clip, trim=1.8cm 1.8cm 1.8cm 1.8cm, width=1\textwidth]{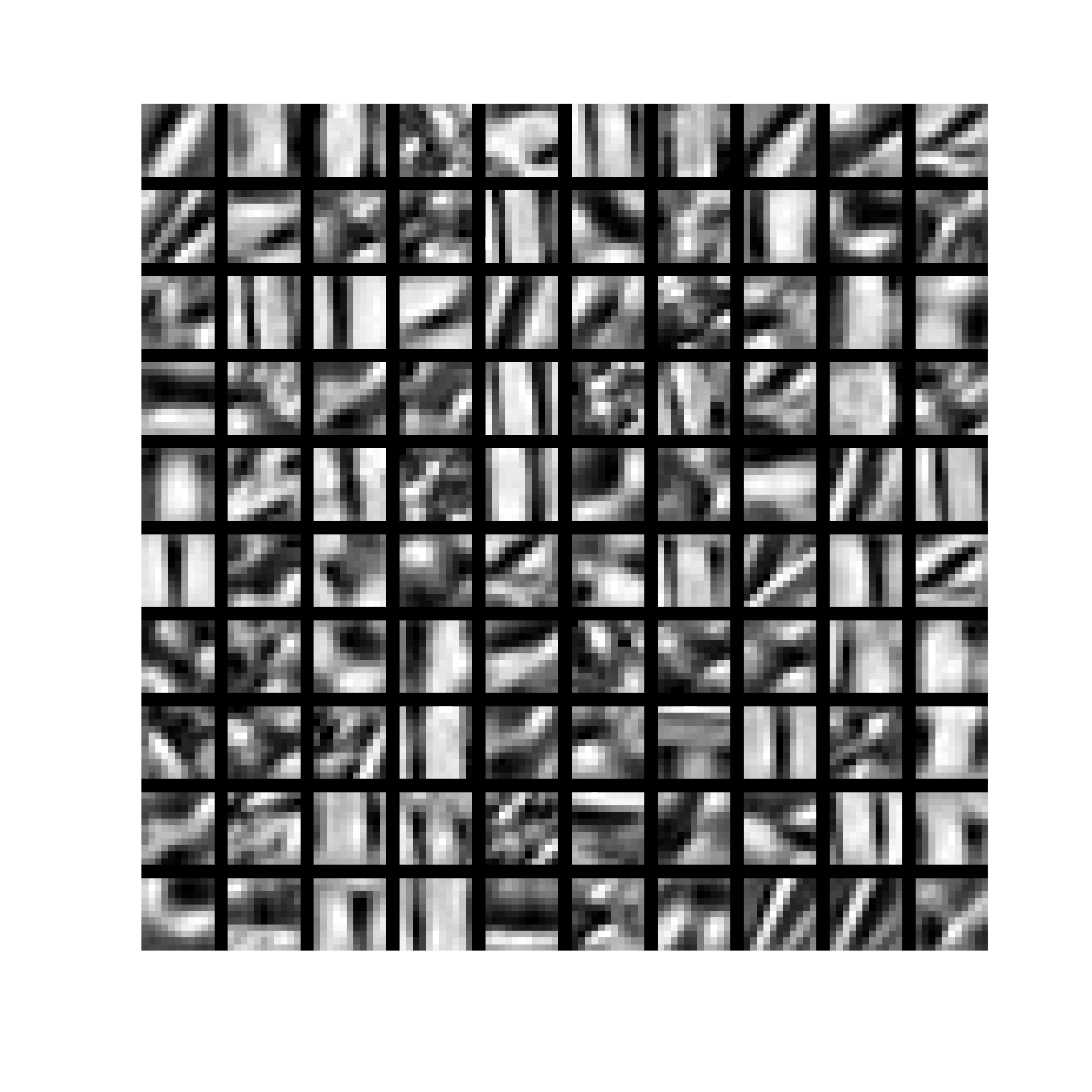}
		\caption{Proposed - Iteration 3.}
	\end{subfigure}
	\begin{subfigure}{0.205\textwidth}
		\centering
		\includegraphics[clip, trim=1.8cm 1.8cm 1.8cm 1.8cm, width=1\textwidth]{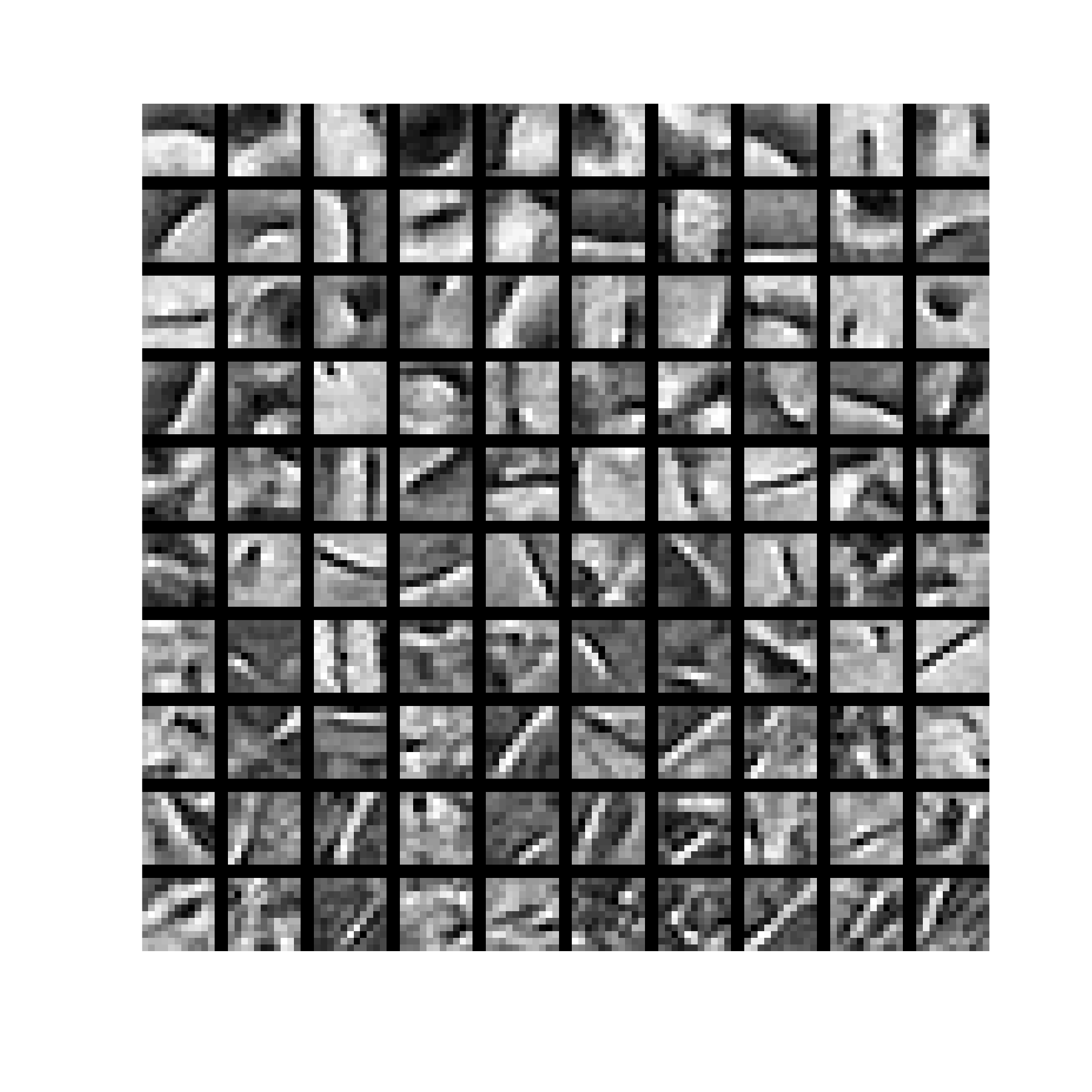}
		\caption{Proposed - Iteration 300.}
	\end{subfigure}
	\\[0.1cm]
	\begin{subfigure}{0.205\textwidth}
		\centering
		\includegraphics[clip, trim=1.8cm 1.8cm 1.8cm 1.8cm, width=1\textwidth]{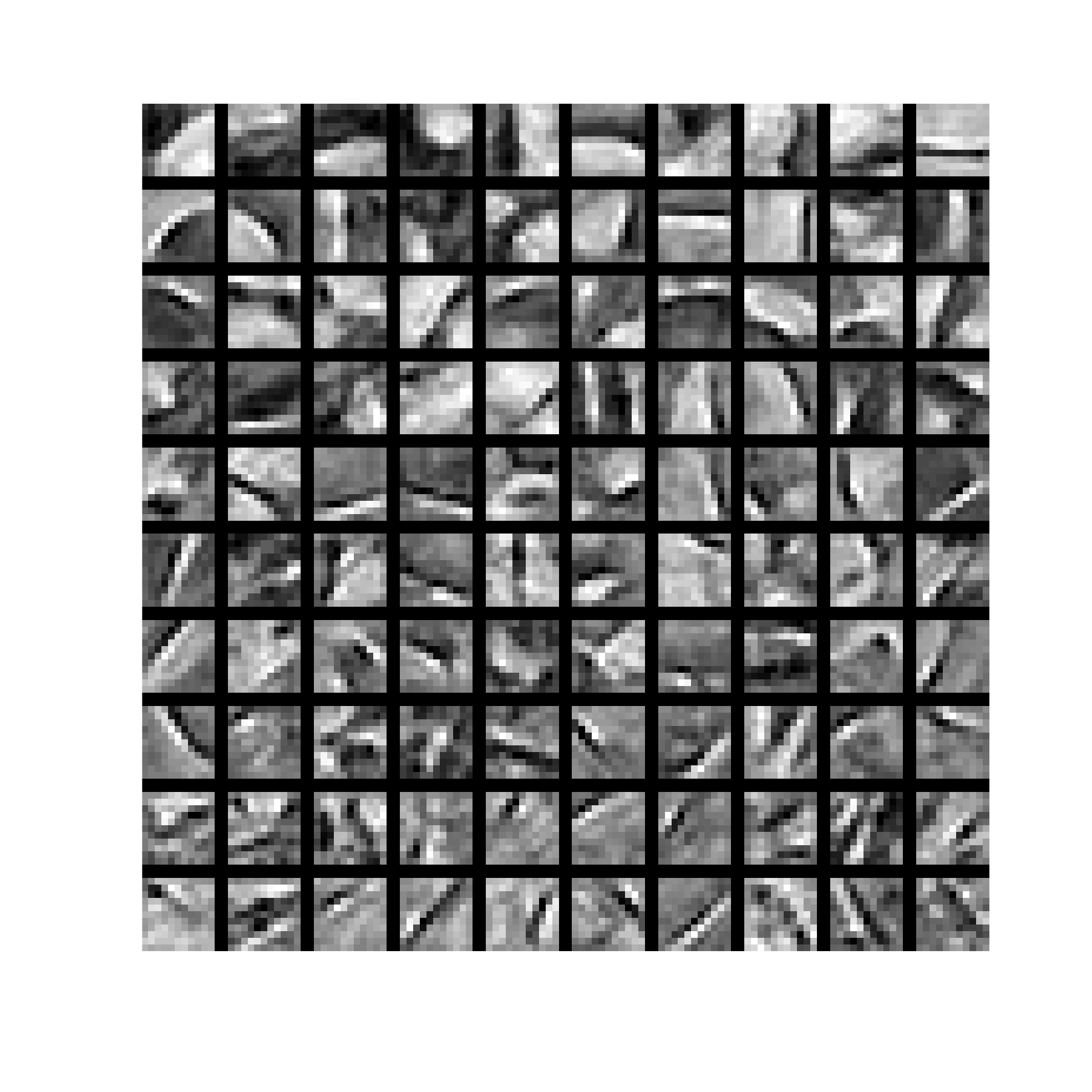}
		\caption{\cite{Heide2015}.}
	\end{subfigure}
	\begin{subfigure}{0.205\textwidth}
		\centering
		\includegraphics[clip, trim=1.8cm 1.8cm 1.8cm 1.8cm, width=1\textwidth]{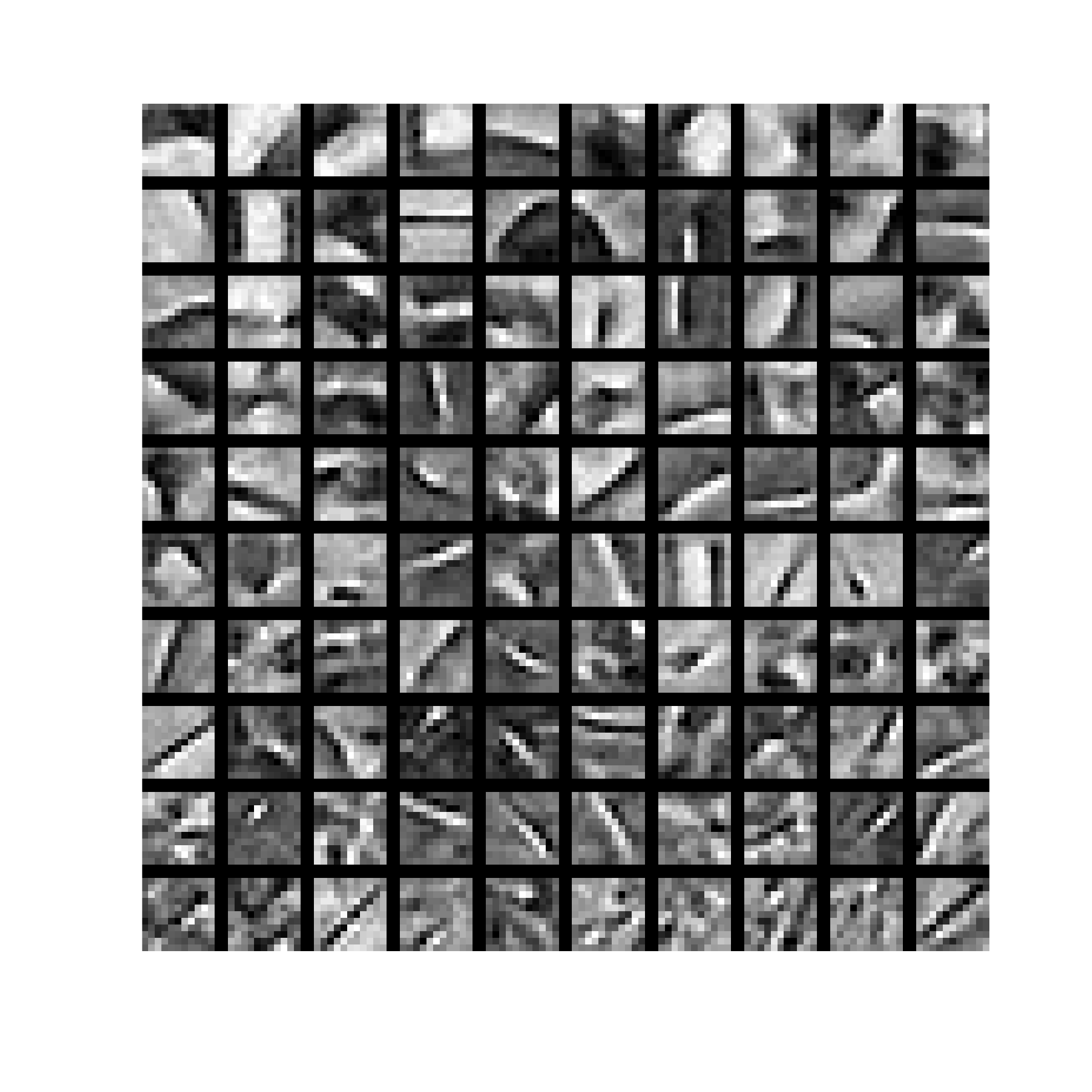}
		\caption{\cite{wohlberg2016boundary}.}
	\end{subfigure}
	\caption{The dictionary obtained after 3 and 300 iterations using the slice-based dictionary learning method. Notice how the atoms become crisper as the iterations progress. For comparison, we present also the result of \cite{Heide2015} and \cite{wohlberg2016boundary}.}
	\label{Fig:single_layer_dictionary}
\end{figure}

\begin{figure}[b]
	\centering
	\includegraphics[width=0.37\textwidth]{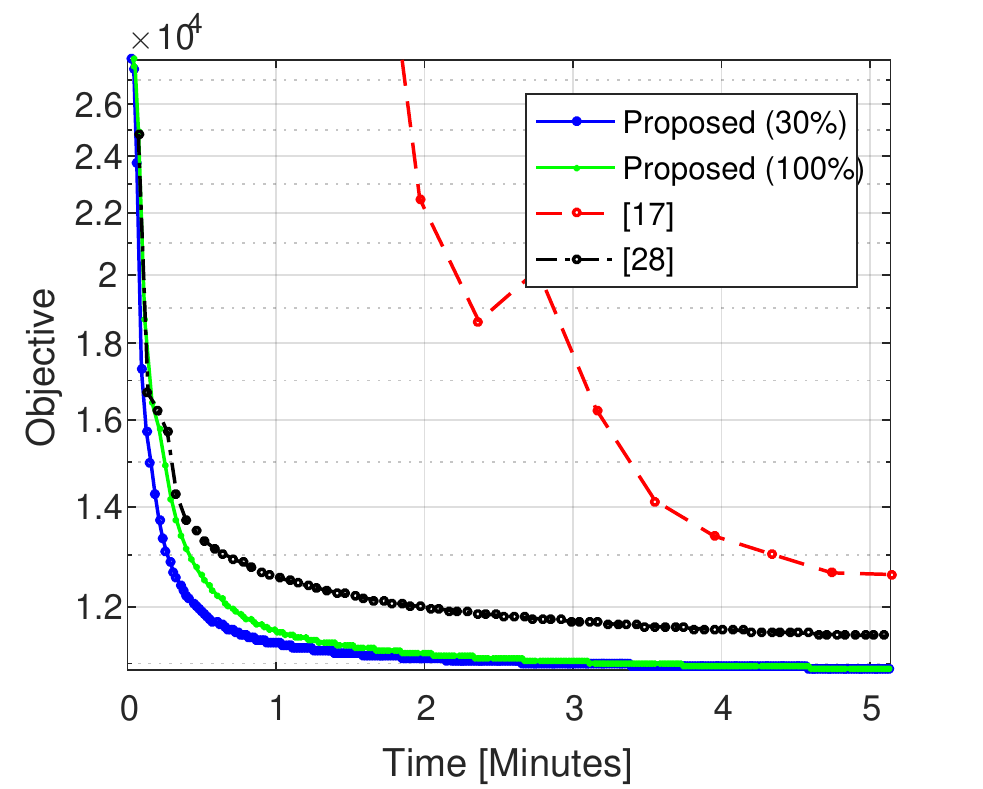}
	\caption{Our method versus the those in \cite{Heide2015} and \cite{wohlberg2016boundary}.}
	\label{Fig:single_layer_obj_vs_time}
\end{figure}

\section{Experiments} \label{Sec:experiments}
We turn to demonstrate our proposed slice-based dictionary learning. Throughout the experiments we use the LARS algorithm \cite{efron2004least} to solve the LASSO problem and the K-SVD \cite{aharon2006rm} for the dictionary learning. The reader should keep in mind, nevertheless, that one could use any other pursuit or dictionary learning algorithm for the respective updates. In all experiments, the number of filters trained are $100$ and they are of size $11 \times 11$.

\subsection{Slice-Based Dictionary Learning}
Following the test setting presented in \cite{Heide2015}, we run our proposed algorithm to solve Equation \eqref{Eq:conv_SC_DL} with $\lambda=1$ on the Fruit dataset \cite{zeiler2010deconvolutional}, which contains ten images. As in \cite{Heide2015}, the images were mean subtracted and contrast normalized. We present in Figure \ref{Fig:single_layer_dictionary} the dictionary obtained after several iterations using our proposed slice-based dictionary learning, and compare it to the result in \cite{Heide2015} and also to the method AVA-AMS in \cite{wohlberg2016boundary}. Note that all three methods handle the boundary conditions, which were discussed in Section \ref{Sec:boundaries}. We compare in Figure \ref{Fig:single_layer_obj_vs_time} the objective of the three algorithms as function of time, showing that our algorithm is more stable and also converges faster. In addition, to demonstrate one of the advantages of our scheme, we train the dictionary on a small subset ($30\%$) of all slices and present the obtained result in the same figure. 

\begin{table*}[t!]
	\centering
	\begin{tabular}{|l|c|c|c|c|c|c|c|c|c|c|c|}
		\hline
		& \small{Barbara}         & \small{Boat}            & \small{House}           & \small{Lena}            & \small{Peppers}        & \small{C.man}      & \small{Couple}          & \small{Finger}     & \small{Hill}            & \small{Man}             & \small{Montage}                \\ \hline \hline
		\small{Heide et al.}                & 11.00          & 10.29          & 10.18          & 11.77          & {9.41} & 9.74          & 11.99          & 15.55          & 10.37          & 11.60          & 15.11                  \\ \hline
		\small{Proposed}                    & {11.67} & {10.33} & {10.56} & {11.92} & 9.18          & {9.95} & {12.25} & {16.04} & {10.66} & {11.84} & {15.40}         \\ \hline
		\small{Image specific}                    & {15.20} & {11.60} & {11.77} & {12.35} & 11.45          & {10.68} & {12.41} & {16.07} & {10.90} & {11.71} & {15.67}         \\ \hline

	\end{tabular}
	\caption{Comparison between the slice-based dictionary learning and the algorithm in \cite{Heide2015} on the task of image inpainting.}\label{Tb:table1}
	\vspace{-0.3cm}
\end{table*}

\begin{figure}[b!]
	\centering
%	\begin{subfigure}{0.13\textwidth}
%		\centering
%		\includegraphics[width=1\textwidth]{original.png}
%		\caption{Original.}
%	\end{subfigure}
%	\begin{subfigure}{0.15\textwidth}
%		\centering
%		\includegraphics[width=1\textwidth]{corrupted.png}
%		\caption{Corrupted.}
%	\end{subfigure}
	\begin{subfigure}{0.15\textwidth}
		\centering
		\includegraphics[width=1\textwidth]{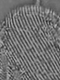}
	\end{subfigure}
	\begin{subfigure}{0.15\textwidth}
		\centering
		\includegraphics[width=1\textwidth]{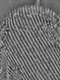}
	\end{subfigure}
	\begin{subfigure}{0.15\textwidth}
		\centering
		\includegraphics[width=1\textwidth]{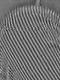}
	\end{subfigure}
	\caption{Visual comparison on a cropped region extracted from the image \textsf{\small{Barbara}}. Left: \cite{Heide2015} (PSNR = 5.22dB). Middle: Ours (PSNR = 6.24dB). Right: Ours with dictionary trained on the corrupted image (PSNR = 12.65dB).}
	\label{Fig:zoomin}
\end{figure}

\begin{figure*}[b!]
	\centering
	\begin{subfigure}{0.12\textwidth}
		\centering
		\includegraphics[width=1\textwidth]{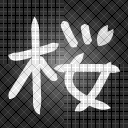}
		\caption{Original.}
	\end{subfigure}
	\begin{subfigure}{0.12\textwidth}
		\centering
		\includegraphics[width=1\textwidth]{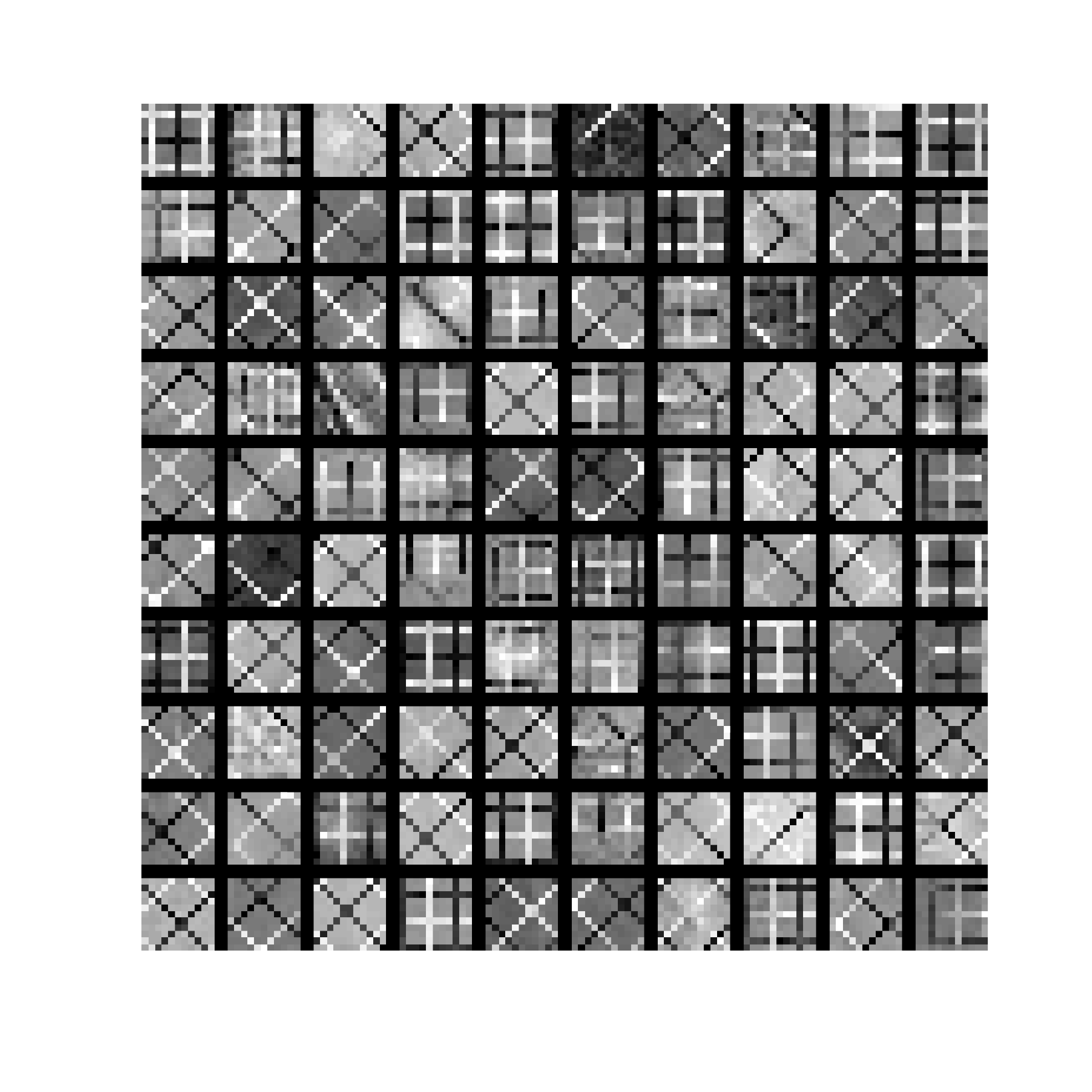}
		\caption{Dictionary.}
	\end{subfigure}
	\hspace{0.5cm}
	\begin{subfigure}{0.12\textwidth}
		\centering
		\includegraphics[width=1\textwidth]{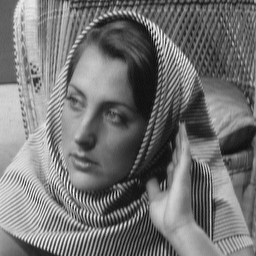}
		\caption{Original.}
	\end{subfigure}
	\begin{subfigure}{0.12\textwidth}
		\centering
		\includegraphics[width=1\textwidth]{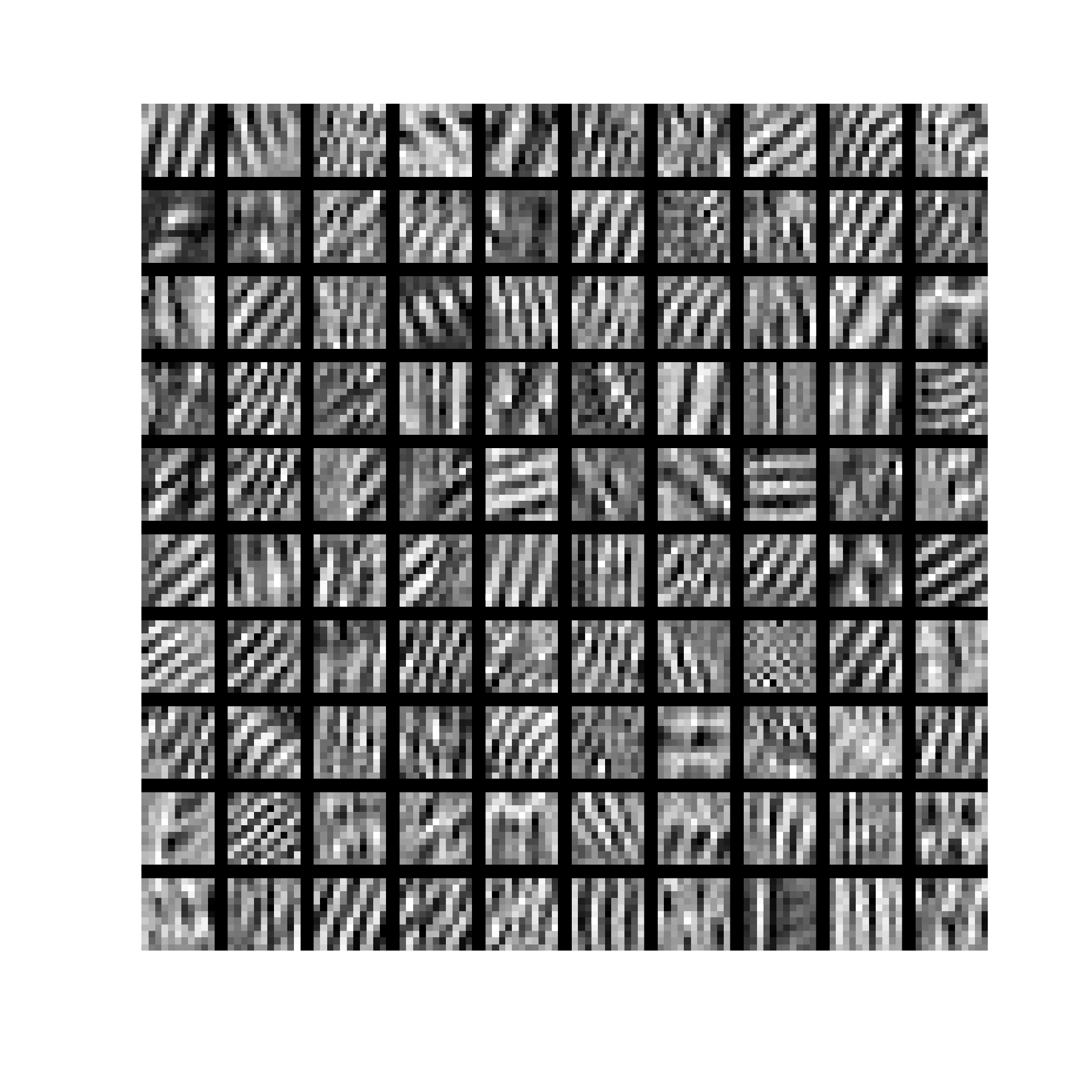}
		\caption{Dictionary.}
	\end{subfigure}
	\hspace{0.5cm}
	\begin{subfigure}{0.12\textwidth}
		\centering
		\includegraphics[width=1\textwidth]{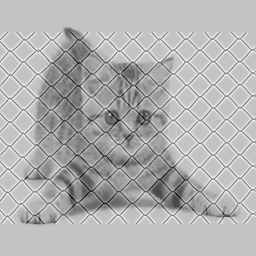}
		\caption{Original.}
	\end{subfigure}
	\begin{subfigure}{0.12\textwidth}
		\centering
		\includegraphics[width=1\textwidth]{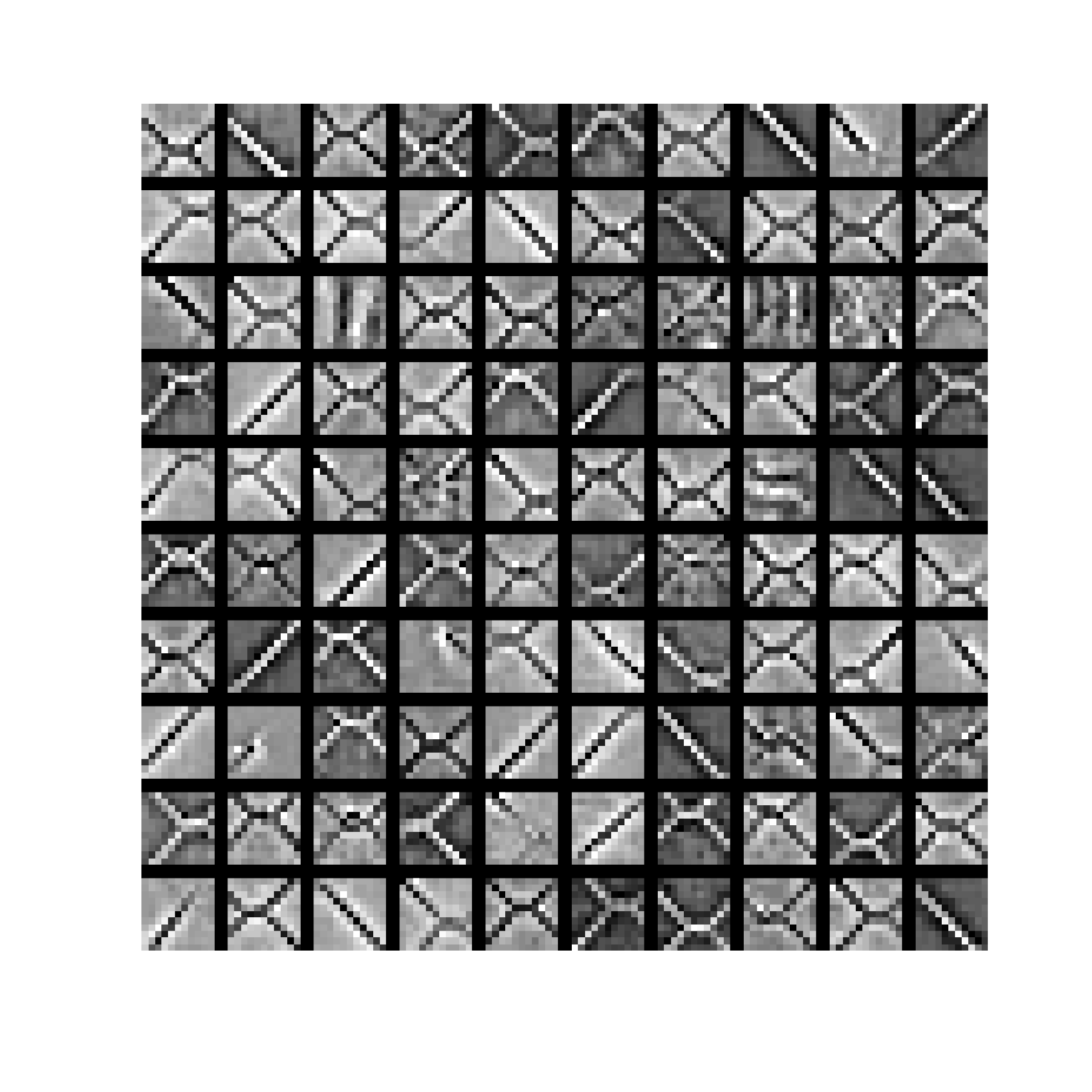}
		\caption{Dictionary.}
	\end{subfigure}
	\\[0.1cm]
	\begin{subfigure}{0.12\textwidth}
		\centering
		\includegraphics[width=1\textwidth]{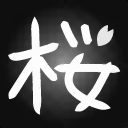}
		\caption{Our cartoon.}
	\end{subfigure}
	\begin{subfigure}{0.12\textwidth}
		\centering
		\includegraphics[width=1\textwidth]{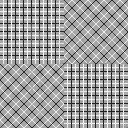}
		\caption{Our texture.}
	\end{subfigure}
	\hspace{0.5cm}
	\begin{subfigure}{0.12\textwidth}
		\centering
		\includegraphics[width=1\textwidth]{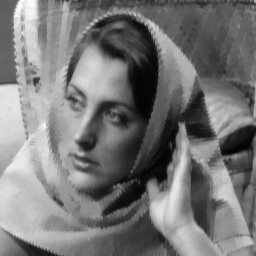}
		\caption{Our cartoon.}
	\end{subfigure}
	\begin{subfigure}{0.12\textwidth}
		\centering
		\includegraphics[width=1\textwidth]{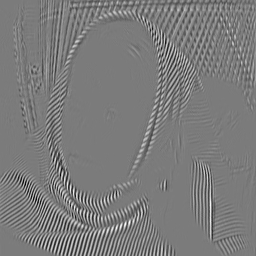}
		\caption{Our texture.}
	\end{subfigure}
	\hspace{0.5cm}
	\begin{subfigure}{0.12\textwidth}
		\centering
		\includegraphics[width=1\textwidth]{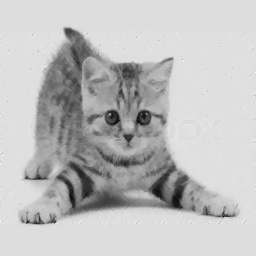}
		\caption{Our cartoon.}
	\end{subfigure}
	\begin{subfigure}{0.12\textwidth}
		\centering
		\includegraphics[width=1\textwidth]{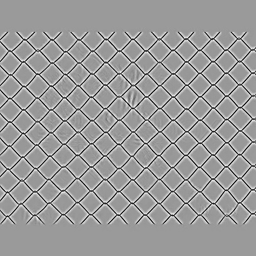}
		\caption{Our texture.}
	\end{subfigure}
	\\[0.1cm]
	\begin{subfigure}{0.12\textwidth}
		\centering
		\includegraphics[width=1\textwidth]{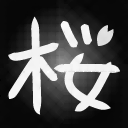}
		\caption{\cite{ono2014cartoon}.}
	\end{subfigure}
	\begin{subfigure}{0.12\textwidth}
		\centering
		\includegraphics[width=1\textwidth]{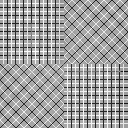}
		\caption{\cite{ono2014cartoon}.}
	\end{subfigure}
	\hspace{0.5cm}
	\begin{subfigure}{0.12\textwidth}
		\centering
		\includegraphics[width=1\textwidth]{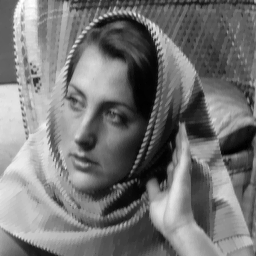}
		\caption{\cite{ono2014cartoon}.}
	\end{subfigure}
	\begin{subfigure}{0.12\textwidth}
		\centering
		\includegraphics[width=1\textwidth]{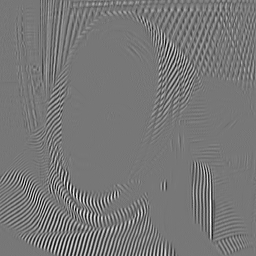}
		\caption{\cite{ono2014cartoon}.}
	\end{subfigure}
	\hspace{0.5cm}
	\begin{subfigure}{0.12\textwidth}
		\centering
		\includegraphics[width=1\textwidth]{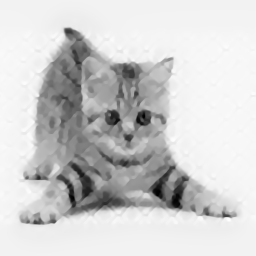}
		\caption{\cite{zhangconvolutional}.}
	\end{subfigure}
	\begin{subfigure}{0.12\textwidth}
		\centering
		\includegraphics[width=1\textwidth]{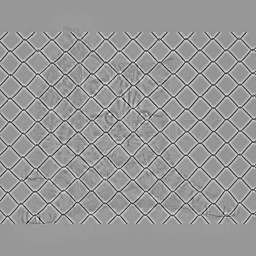}
		\caption{\cite{zhangconvolutional}.}
	\end{subfigure}
	\caption{Texture and cartoon separation for the images \textsf{\small{Sakura}}, \textsf{\small{Barbara}} and \textsf{\small{Cat}}.}
	\label{Fig:separation}
\end{figure*}

\subsection{Image Inpainting}
We turn to test our proposed algorithm on the task of image inpainting, as described in Section \ref{Sec:inpainting}. We follow the experimental setting presented in \cite{Heide2015} and compare to their state-of-the-art method using their publicly available code. The dictionaries employed in both approaches are trained on the Fruit dataset, as described in the previous subsection (see Figure \ref{Fig:single_layer_dictionary}). For a fair comparison, in the inference stage, we tuned the parameter $\lambda$ for both approaches. Table \ref{Tb:table1} presents the results in terms of peak signal-to-noise ratio (PSNR) on a set of publicly available standard test images, showing our method leads to quantitatively better results\footnote{The PSNR is computed as $20\log ( \sqrt{N} / \| \X - \hat{\X} \|_2 )$, where $\X$ and $\hat{\X}$ are the original and restored images. Since the images are normalized, the range of the PSNR values is non-standard.}. Figure \ref{Fig:zoomin} compares the two visually, showing our method also leads to better qualitative results.

A common strategy in image restoration is to train the dictionary on the corrupted image itself, as shown in \cite{Elad2006}, as opposed to employing a dictionary trained on a separate collection of images. The algorithm presented in Section \ref{Sec:inpainting} can be easily adapted to this framework by updating the local dictionary on the slices obtained at every iteration. To exemplify the benefits of this, we include the results\footnote{A comparison with the method of \cite{Heide2015} was not possible in this case, as their implementation cannot handle training a dictionary on standard-sized images.} obtained by using this approach in Table \ref{Tb:table1} and Figure \ref{Fig:zoomin}.

%One should note that once the dictionaries are fixed, the sparse pursuit problem is convex, and thus the different algorithms should give the exact same result. As such, this experiment indicates that the dictionary obtained by our method leads to better performance, standing in agreement with the lower objective obtained by our dictionary learning algorithm (see Figure \ref{Fig:single_layer_obj_vs_time}).

\begin{figure}[t!]
	\centering
	\begin{subfigure}{0.2\textwidth}
		\centering
		\includegraphics[width=1\textwidth]{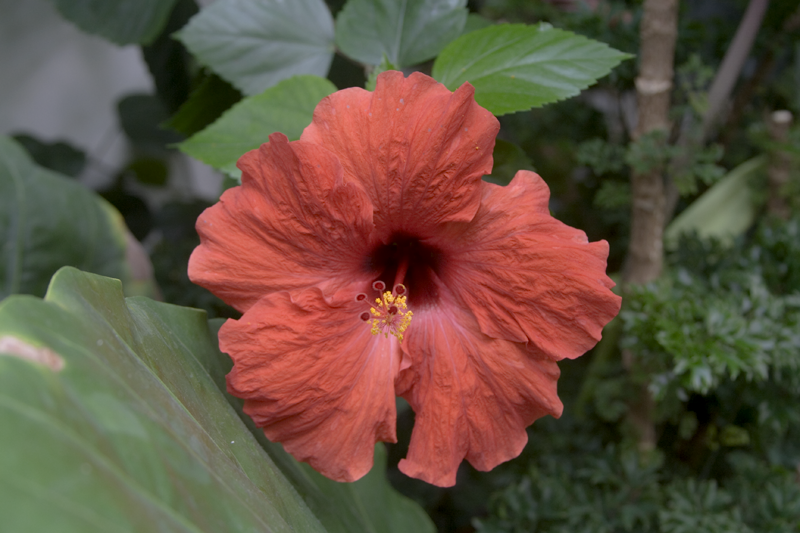}
		\caption{Original image.}
	\end{subfigure}
	\begin{subfigure}{0.2\textwidth}
		\centering
		\includegraphics[width=1\textwidth]{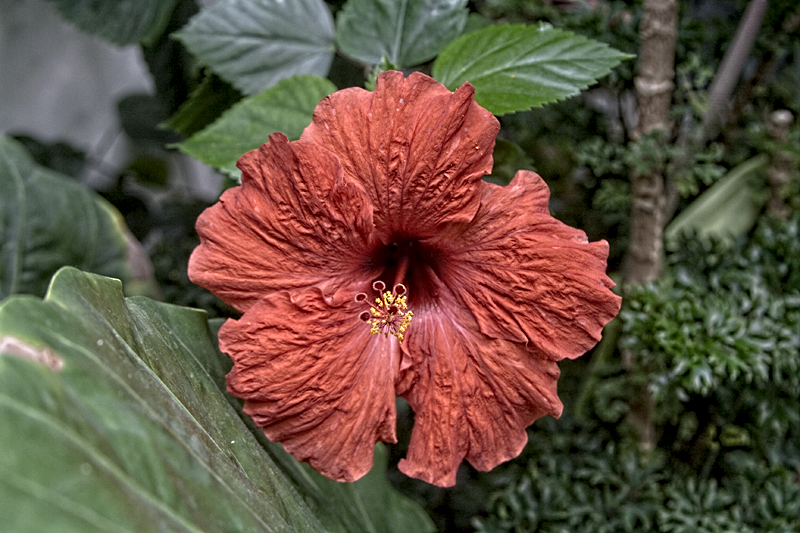}
		\caption{Enhanced output.}
	\end{subfigure}
	\caption{Enhancement of the image \textsf{\small{Flower}} via cartoon-texture separation.}
	\label{Fig:enhancement}
	\vspace{-0.5cm}
\end{figure}

\subsection{Texture and Cartoon Separation}
We conclude by applying our proposed slice-based dictionary learning algorithm to the task of texture and cartoon separation. The TV denoiser used in the following experiments is the publicly available software of \cite{chan2011augmented}. We run our method on the synthetic image \textsf{\small{Sakura}} and a portion extracted from \textsf{\small{Barbara}}, both taken from \cite{ono2014cartoon}, and on the image \textsf{\small{Cat}}, originally from \cite{zhangconvolutional}. For each of these, we compare with the corresponding methods. We present the results of all three experiments in Figure \ref{Fig:separation}, together with the trained dictionaries. Lastly, as an application for our texture separation algorithm, we enhance the image \textsf{\small{Flower}} by multiplying its texture component by a scalar factor (greater than one) and combining the result with the original image. We treat the colored image by transforming it to the Lab color space, manipulating  the L channel, and finally transforming the result back to the original domain. The original image and the obtained result are depicted in Figure \ref{Fig:enhancement}. One can observe that our approach does not suffer from halos, gradient reversals or other common enhancement artifacts.

\vspace{-0.1cm}

\section{Conclusion} \label{Sec:conclusion}
In this work we proposed the slice-based dictionary learning algorithm. Our method employs standard patch-based tools from the realm of sparsity to solve the global CSC problem. We have shown the relation between our method and the patch-averaging paradigm, clarifying the main differences between the two: (i) the migration from patches to the simpler entities called slices, and (ii) the application of a local Laplacian that results in a global consensus. Finally, we illustrated the advantages of the proposed algorithm in a series of applications and compared it to related state-of-the-art methods.

\newpage

%\section{Acknowledgments}
%The research leading to these results has received funding from the European Research Council under European Union's Seventh Framework Programme, ERC Grant agreement no. 320649.

{\small
\bibliographystyle{ieee}
\bibliography{egbib}
}

\end{document}